%% file: main.tex
\documentclass[10pt,twocolumn,letterpaper,pagebackref,breaklinks,colorlinks,allcolors=cvprblue]{article}

\usepackage{cvpr}              
\definecolor{cvprblue}{rgb}{0.21,0.49,0.74}
\usepackage{hyperref}
\usepackage{adjustbox}
\usepackage{tabularx}
\usepackage{makecell}
\usepackage{subcaption}
\usepackage{pifont}
\usepackage{stmaryrd}
\usepackage{multicol}

\title{A Large-Scale Vision-Language Dataset Derived from Open Scientific Literature to Advance Biomedical Generalist AI}

\author{Alejandro Lozano*$^1$
\and
Min Woo Sun*$^1$\\
\and
James Burgess*$^1$ \\
\and
Jeffrey J. Nirschl$^1$\\
\and
Christopher  Polzak$^1$ \\
\and
Yuhui Zhang$^1$\\
\and
Liangyu Chen$^1$ \\
\and
Jeffrey Gu$^1$ \\
\and
Ivan Lopez$^1$ \\
\and
Josiah Aklilu$^1$ \\
\and
Anita Rau$^1$ \\
\and
Austin Wolfgang Katzer$^1$ \\
\and
Collin Chiu$^1$ \\
\and
Orr Zohar$^1$ \\
\and
Xiaohan Wang$^1$ \\
\and
Alfred Seunghoon Song$^1$\\
\and
Chia-Chun Chiang$^2$ \\
\and
Robert Tibshirani$^1$\\
\and
Serena Yeung-Levy$^1$ \\ \\
$^1$Stanford University \hspace{1em} $^2$Mayo Clinic
}

\newcommand{\dataset}{BIOMEDICA}

\newcommand{\modelMed}{BMC-CLIP}

\begin{document}
\input{sec/0_abstract}

\input{sec/1_intro}

\input{sec/5_results}

{
    \small
    \bibliographystyle{ieeenat_fullname}
    \bibliography{main}
}

\input{sec/X_suppl}

\end{document}

%% file: sec/0_abstract.tex
\twocolumn[{%
\maketitle
\renewcommand\twocolumn[1][]{#1}%
\vspace{-1em}
\begin{abstract}

Despite the excitement behind biomedical artificial intelligence (AI), access to high-quality, diverse, and large-scale data -- the foundation for modern AI systems -- is still a bottleneck to unlocking its full potential. To address this gap, we introduce \dataset\, an open-source dataset derived from the PubMed Central Open Access subset, containing over 6 million scientific articles and 24 million image-text pairs, along with 27 metadata fields, including expert human annotations. To overcome the challenges of accessing our large-scale dataset, we offer a web platform with tools that enable both targeted content retrieval and on-demand data access without downloading the entire dataset, facilitating seamless integration with AI systems. We demonstrate the utility of the \dataset\ dataset by building embedding models, chat-style models, and  retrieval-augmented chat agents. Notably, all our AI models surpass previous open systems in their respective categories, underscoring the critical role of diverse, high-quality, and large-scale  biomedical data.

\end{abstract}
}]

%% file: sec/1_intro.tex
\label{sec:intro}
\begin{figure*}[h]
    \centering
    \includegraphics[width=0.85\textwidth]{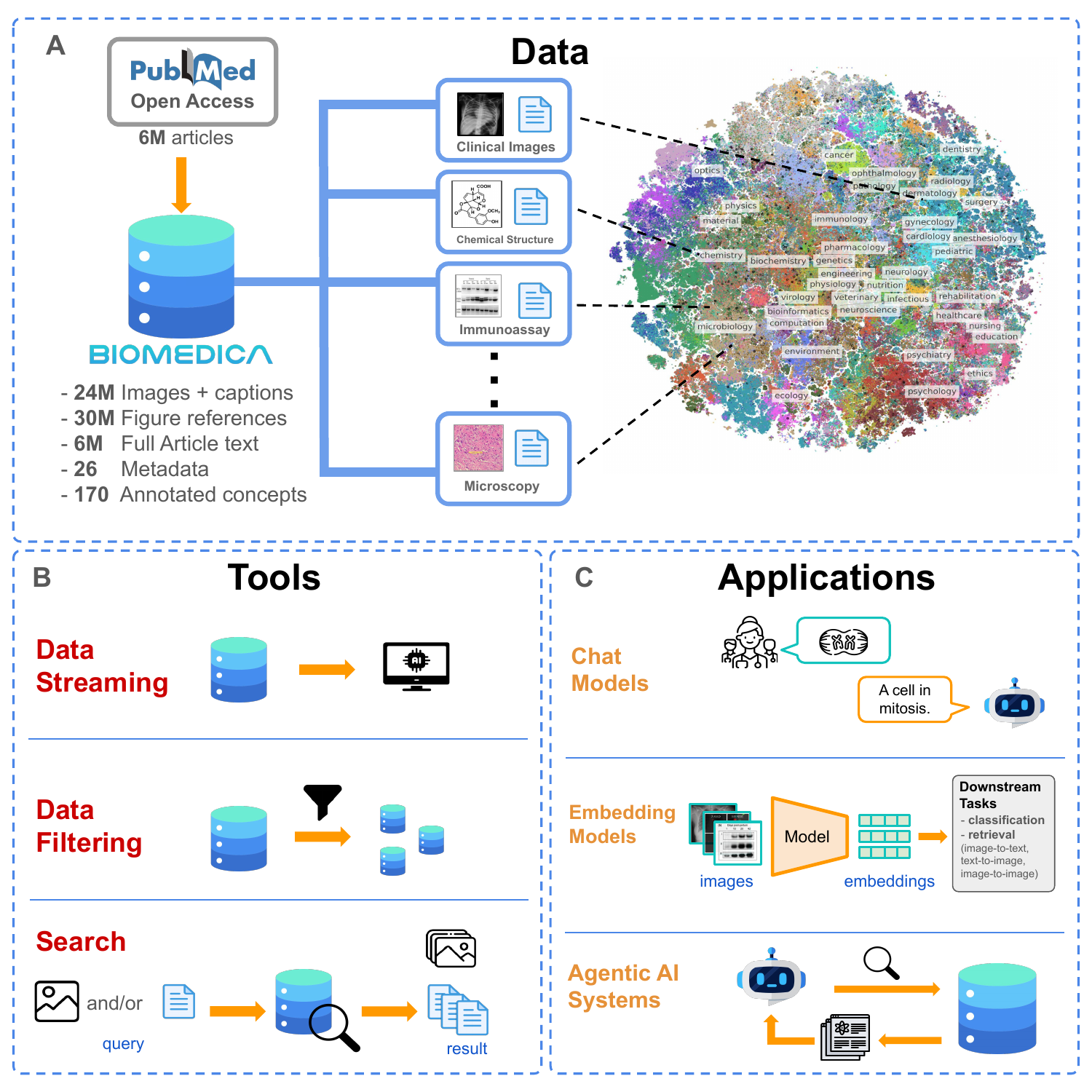}
    \vspace{-0.7em}
    \caption{
    Overview of the \dataset\ dataset, tools for accessibility, and its applications. (A) Overlap of \dataset\  with the Landscape
of Biomedical Research \cite{gonzalez2024landscape}: The dataset comprises 6 million open-access articles, 24 million image-caption pairs, and 30 million in-line references, spanning diverse biomedical domains such as clinical radiology and pathology images, research microscopy, immunoassays, chemical structures, among other scientific images. (B) To facilitate AI model development and inference, we offer data streaming, filtering, and the \dataset\ Index. Streaming enables efficient training without the need for extensive local storage. Data filtering allows users to create domain-specific subsets of the data. The \dataset\ Index supports multi-modal retrieval-based applications. (C) The \dataset\ dataset enables diverse biomedical applications, including chat models, embedding models, and agentic systems.}
    \label{Fiugre1}
\end{figure*}

\begin{figure*}[ht]
  \centering
  \includegraphics[width=1\textwidth]{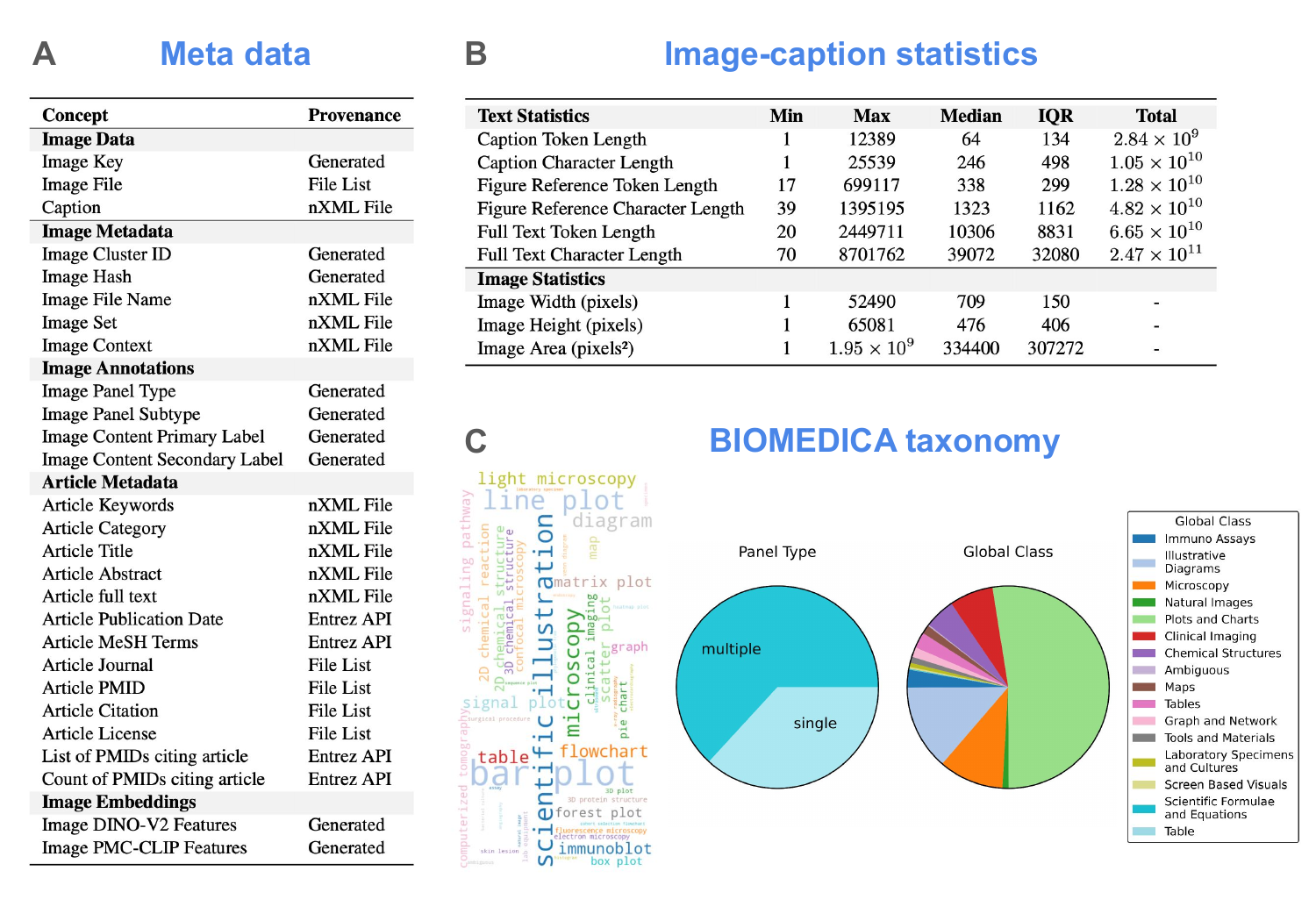}
  \vspace{-3.5em}
  \caption{Overview of the \dataset\ dataset statistics and annotations: (A) List of metadata fields provided in the dataset, along with their respective sources of provenance. (B) Summary statistics for the dataset's text tokens, characters, and image dimensions. Image statistics include width, height, and area in pixels. (C) Distribution of image annotations in the dataset. The word cloud on the left visualizes the most common concept labels assigned during annotation. The left pie chart shows the proportion of images categorized as single-panel or multi-panel. The right pie chart presents the distribution of global image classes, representing various biomedical categories.}
\end{figure*}

In recent years,  foundation models (FMs) -- large deep learning systems trained on massive datasets -- have demonstrated increasing expert-level performance across a range of biomedical tasks such as diagnosis and prognosis from clinical images, patient report generation, medical conversation, and scientific literature summarization~\cite{moor2023foundation,saab2024capabilities,burgess2025microvqa}. 
This flexibility and reusability of FMs is mainly driven by large and diverse datasets~\cite{udandarao2024no} -- a process known as data scaling.
However, access to such diverse datasets throughout medicine and biology is limited, as existing data artifacts are often private or difficult to access, creating a significant bottleneck to achieving progress comparable to that seen in the general domain with systems like GPT-4~\cite{achiam2023gpt}.

Through original research articles, systematic reviews, meta-analyses, case studies, clinical trials, and commentaries: scientific biomedical literature provides an ever-expanding, highly curated multimodal resource encompassing the knowledge of specialized professionals -- reflecting rigorously supported medical and biological evidence.
Naturally, open-source biomedical literature offers an unparalleled resource for constructing comprehensive and diverse datasets at scale. While prior efforts have leveraged this resource \cite{pelka2018radiology, lin2023pmc}, the existing open-source literature-based datasets have several limitations. First, they include images and captions but not the article text, thereby discarding rich training data. Second, they lack bibliographic and image metadata, which are crucial for modern pretraining techniques (e.g. data balancing and filtering).
Finally, they are pre-filtered to diagnostics imaging modalities within radiology and pathology, overlooking the vast breadth of complementary information available in fields such as cancer biology, genetics, and pharmacogenomics, which provide key insights into disease mechanisms and are increasingly vital to precision medicine.

To democratize access to open-source scientific data across the vast landscape of biomedical research, we extract and annotate PubMed Central Open Access to introduce the Biomedical Image-Caption Archive (\dataset) — a comprehensive dataset containing image-caption pairs, image-references pairs,  full-text scientific articles,  metadata, and human-derived annotations. Given the scale of the dataset, accessibility presents practical engineering challenges. Thus, we host \dataset\ on a web server and offer tools for efficient access -- optimizing both high-volume access for AI model training and fast, on-demand search for agentic applications. We demonstrate the utility of \dataset\ by building multiple AI applications: embedding models, chat-style models, and retrieval-augmented (RAG) chat systems. Notably, all models surpass previous open systems in their respective model categories, underscoring the importance of high-quality, diverse, and large-scale datasets.

%% file: sec/5_results.tex
\section*{Results and Discussion}

\dataset\ is a large-scale, diverse, and multimodal dataset for building biomedical AI systems. It consists of 6 million scientific articles and 24 million image-caption pairs. 
The articles cover open-access research literature throughout all of medicine and life sciences. Images also span diverse categories including clinical imaging, microscopy, chemical structures, immunoassays, illustrative diagrams, tables, and maps (Figure \ref{Fiugre1}A).

 \dataset\ is extensively annotated, with each instance (articles, images, and captions) containing both article- and image-level metadata and expert-derived annotations. Article-level metadata includes bibliographic details such as PMID, publication date, citation, journal, license, title, abstract, MeSH terms, keywords, and references to citing articles. Image metadata is collected through expert-derived annotations,  which classify image content into global and local categories, as well as panel-type annotations (e.g., single-panel or multi-panel). These additional annotations enable leveraging training strategies such as concept balancing and data filtering, and facilitate retrieving data on demand based on personalized queries.

\begin{figure*}[ht]
  \centering
  \includegraphics[width=1\textwidth]{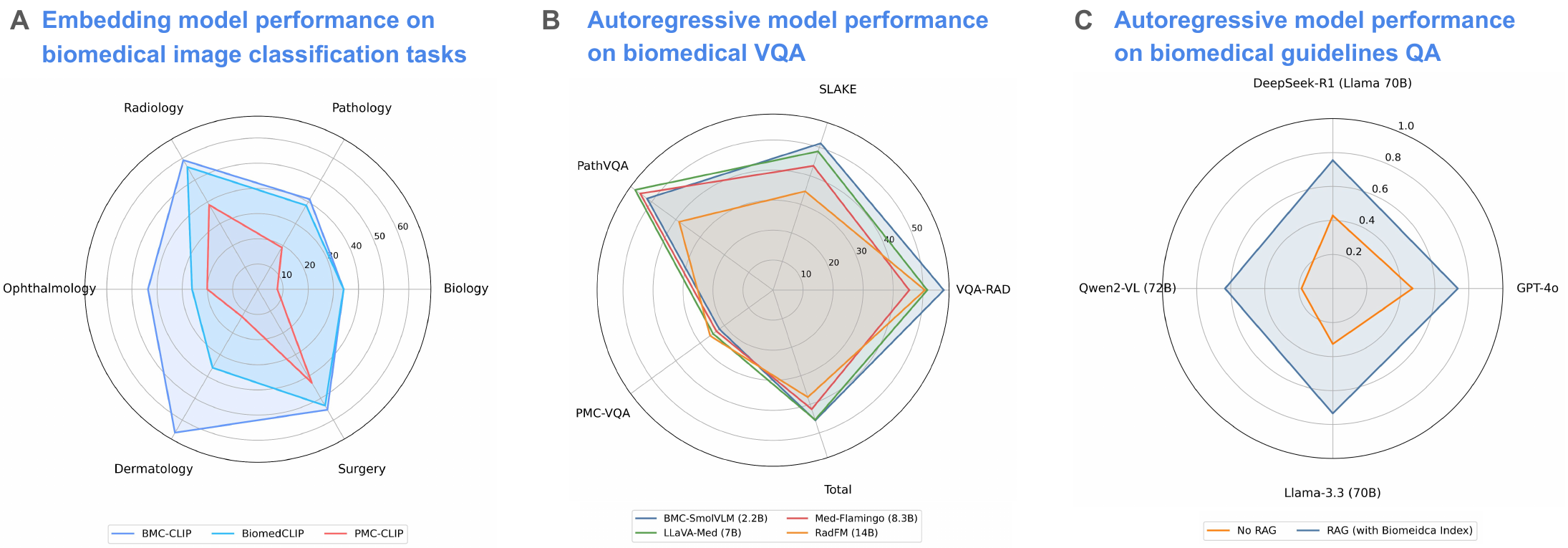}
  \vspace{-1.8em}
  \caption{\dataset\ enables state of the art performance across multiple applications. (A) Multimodal embedding model performance on biomedical image classification tasks. (B) Autoregressive model performance on biomedical VQA tasks (results for other previous models are obtained from \cite{zhang2023huatuogpt}).
 (C) Autoregressive model performance on biomedical guidelines QA across four LLMs with and without retrieval augmentation using the \dataset\ Index.}
\label{Figure_eval}
\end{figure*}

The large-scale data provided by \dataset\ introduces a significant challenge for most users: it is too resource-intensive to store and manage. We therefore host all data artifacts on a Hugging Face, and provide tools for efficient access (Figure \ref{Fiugre1}B). The dataset can be continuously streamed,  allowing users to train ML models without storing any data locally. 
Additionally, the streaming dataset can be filtered based on metadata or annotations, enabling the creation of tailored subsets for specialist models. Common image-type subsets, including histopathology images and medical diagrams, have also been made available on Hugging Face. In addition to training, many AI applications require fast data \textit{search}: for example, retrieval-augmented chat systems \cite{lozano2023clinfo}, or more autonomous AI agent systems \cite{narayanan2024aviary}. We therefore release an application programming interface (API) for search over articles, captions, and images, where the query can be text, images, or both. We implement this as a hybrid `vector-text' database called the \dataset\ Index.

\vspace{1em}

The scale and comprehensive content of \dataset\ establish it as a foundational resource for developing a wide range of AI systems. To demonstrate its utility, we create three demonstrator applications: an embedding model, a chat model, and a retrieval-augmented chat system. We curate widely-used multimodal biomedical benchmarks to evaluate and compare the performance of our embedding and chat models against previous work.


\vspace{1em}

\noindent \textbf{Embedding models trained on the \dataset\ dataset lead to the better multimodal representations.} By learning shared representations across diverse data types, joint vision-language modeling through contrastive learning has emerged as a powerful approach to improve image classification and multimodal retrieval \cite{radford2021learning}. Within biomedical applications, integrating multiple imaging modalities enables a more comprehensive and holistic view of diseases and patient states. To this end, we leverage annotations and metadata from the \dataset\ dataset to filter out non-biomedical images (e.g., plots and tables) and pretrain a contrastive vision-language model,  \modelMed. Our systematic evaluations across 41 datasets show that \modelMed\ outperforms prior work (Figure \ref{Figure_eval}A). Compared to PMC-CLIP \cite{eslami2023pubmedclip}, \modelMed\ achieves a +24.67\% improvement in biomedical image classification, +36.91\% in image-to-text recall@100, and +34.6\% in text-to-image recall@100. Similarly, \modelMed\ surpasses BiomedCLIP \cite{zhang2023biomedclip} in 8 of 10 classification subsets, with an average gain of 6.56\%, and demonstrates marginally better retrieval performance (Table \ref{table:retrieval_performance}).

\vspace{1em}

\noindent \textbf{Vision-language autoregressive models trained on the \dataset\ dataset achieve competitive performance at a fraction of the cost.} By leveraging text as an interface, vision-language autoregressive models enable intuitive, barrier-free AI interaction \cite{saab2024capabilities, li2024llava}. Thus, through natural language queries, these chat-like systems can perform tasks such as visual question answering, image captioning, object detection, and referral segmentation -- key tasks that translate directly to clinical and biological applications. To this end, we use a small subset of the \dataset\ dataset to create a 2M alignment dataset and collect the training sets for VQA-RAD \cite{lau2018dataset}, SLAKE \cite{liu2021slake}, PathVQA \cite{he2020pathvqa}, and PMC-VQA \cite{zhang2023pmc} to add instruction-following data (following the same recipe as LLaVA-Med \cite{li2023llava}). We then fine-tune SmolVLM \cite{marafioti2025smolvlm}, a 2.2B parameter model, to create BMC-SmolVLM and evaluate our system with four vision question-answering tasks. With only 2.2B parameters, our system achieves similar or better performance than previously small models (less than 15B), even outperforming some larger models (Figure \ref{Figure_eval}B). Compared to RADFM (13M), it shows an improvement of 8.09\%, and compared to LMed-Flamingo (8M) \cite{moor2023med}, it demonstrates a +3.89\% improvement. Notably, BMC-SmolVLM achieves performance comparable to LLaVA-Med \cite{li2024llava}, a 7B model.

\vspace{1em}

\noindent \textbf{The \dataset\ Index enables AI agentic systems to answer medical guideline-derived questions.} Agentic systems can potentially assist medical practitioners in their daily workflows, from virtual tumor boards to virtual case conferences. To this end, we use our previously developed embedding models and the \dataset\ index to create a retrieval-augmented  AI agent (BMC-agent). Our system is the first of its kind, capable of querying similar images, captions, and full-text articles using images, text or both to query data across the entire dataset efficiently with an average 123.89ms $\pm$ 4.07ms latency (Figure \ref{query_time}). Furthermore, BMC-agent allows easy  integration of any public or closed VLM/LLM. To evaluate BMC-agent, we curated a clinician-verified dataset of 50 questions derived from  neurology,  molecular pathology, and pharmacogenomics guidelines published up to January 2023 (before the GPT-4o knowledge cut-off in April 2023). We test our AI agent using four different LLMs/VLMs: DeepSeek-R1 (Llama-based 70B)\cite{deepseekai2025}, Qwen2-VL (72B) \cite{wang2024qwen2}, GPT-4o \cite{achiam2023gpt}, and Llama-3.3 (70B) \cite{touvron2023llama}. Our evaluations show that incorporating the \dataset\ index - which allows agents to search and synthesize full-text articles with relevant information - improves performance by an average of 36.22\%  across all evaluated models (Figure \ref{Figure_eval}C).

\section*{Conclusion} 

\dataset\ provides an unprecedented, open-source resource that advances biomedical AI research by democratizing access to diverse, densely annotated, and large-scale multimodal scientific data. It offers tools that facilitate high-volume access, efficient multimodal search, and filtering. Our results demonstrate that leveraging \dataset\ enables the development of high-performance, fully open-source multimodal systems. We believe these features will play a pivotal role in shaping the future of biomedical AI, driving scientific discovery, and advancing personalized healthcare.

\section*{Acknowledgments}

We would like to acknowledge Stanford Data Science, the NVIDIA Academic Grant Program, and AWS for providing computing resources for this project.

This research was supported by NIH grants (NIH \# P30AG066515 to JJN), a Hoffman-Yee Research Grant to SYL, the Arc Institute Graduate Fellowship to AL, the Stanford Data Science Scholars program to MW and the Quad Fellowship to JB. SYL is a Chan Zuckerberg Biohub – San Francisco Investigator.

We thank Laura Bravo, and Maximillian Schuessler for their invaluable discussions. We also thank Daniel van Strien, Matthew Carrigan, and Omar Sanseviero from Hugging Face for their invaluable assistance with data upload and design planning on the Hugging Face platform.

\section*{Author contributions}
AL, MW, JB, and SYL designed the study. MW collected and standardized the training dataset. JN, AS, and AL developed the topic taxonomy. IL, AWK, CC, JN, AS, AL, and MW annotated the dataset. AR, JG, JA, AL, MW, JN, and JB standardized the evaluation datasets. AL and MW trained the contrastive and auto-regressive multimodal models. OZ supervised the auto-regressive multimodal model training. CP developed the index API. AL designed the retrieval-augmented agent. JN guided evaluation selection. JN, CC, and AL curated clinical benchmarks for the retrieval-augmented agent. SYL supervised the study. All authors provided feedback and participated in the writing process.

%% file: sec/X_suppl.tex
\clearpage
\setcounter{page}{1}

\section*{Methods}

\subsection*{Data}
\subsubsection*{Dataset collection}
We downloaded all media files—including images and full article text—from PMC Open Access using the FTP service provided by the National Center for Biotechnology Information (NCBI): \url{https://ftp.ncbi.nlm.nih.gov/pub/pmc/oa_package}. Each article’s media files are stored on the server as a compressed \verb|tar.gz| file, and the file paths are listed in a CSV file called ``file list''. We iterated through this file list and downloaded media files for 6,042,494 articles, retaining the nXML file containing the full article text and the image files. We then used a custom parser to extract figure captions and inline references from the nXML files corresponding to the available images. Additionally, we collected metadata from three  sources: the Entrez API, the file list, and the nXML file. Lastly, we store the data and metadata for a given article as a dictionary and aggregated  multiple articles into a list that are subsequently saved as a JSONL file. The data collection process  took 81 days (from July 2 to September 20, 2024) and required 30TB of storage.

\subsubsection*{Dataset serialization}
We reformatted the dataset from an article-level representation (where each entry contains multiple images) to an image-caption pair level--ensuring that each data element corresponds to a single image, its associated caption, and its metadata. We additionally included the license information for each article to comply with PMC Open Access data usage agreements and their respective copyright terms. Lastly, we serialized the dataset into the WebDataset format: tar files, each containing 10,000 image-caption pairs. This format enables efficient data loading and streaming, allowing models to process image-caption pairs on demand without requiring the entire dataset to be loaded into memory or stored locally.

 \subsubsection*{Dataset Annotation}
To annotate images without introducing a topic prior, we designed an expert-guided framework to cluster and categorize related concepts represented in PMC-OA. The pipeline consists of the following steps:

\begin{enumerate}
    \item We created an initial concept taxonomy based on biomedical ontologies. Using unsupervised clustering on embedded image features (via DINOv2 \cite{oquab2023dinov2}), we grouped similar images. Dimensionality reduction was performed using PCA (25 principal components), followed by K-means clustering (K=2000) to form image groups.
    \item A team of 7 annotators (clinicians and scientists) annotated the clusters based on 30 randomly sampled images from each cluster. Annotators assigned one or more global and local concept labels from the hierarchical taxonomy. The taxonomy was iteratively refined and expanded by the annotators.
    \item We propagated the cluster label to all the images in each cluster, effectively annotating all the images in the dataset.
\end{enumerate}

Using this pipeline, we were able to annotate 24M images using only 24 annotation hours across all 7 experts.

\subsubsection*{Dataset to Index Conversion}
We restructured the dataset into an index to enable multimodal search. To this end, we serialized the images, captions, and full-text articles in two forms:
\begin{enumerate}
\item First, we represent the images and captions as vector embeddings obtained from foundation models (our text embedding models and  DINOv2) and store them in a vector database (ChromaDB). Once stored, the data can be queried by embedding prompt text or images and searching by vector similarity.

\item Second, we build BM25s \cite{bm25s} indices for the captions and full-text articles, which represent items by the word frequencies of their contents for keyword-based search. For efficiency, we restrict our BM25s query vocabulary to frequent tokens (words) appearing in $\geq $ 5 documents. We utilize caching, chunking, and sparse matrices to pre-compute and pre-load all BM25s word frequency scores at once.
\end{enumerate}

We optimize our vector databases and BM25s indices by storing only the minimal metadata needed to uniquely identify items. Full data and metadata for relevant items are retrieved on demand from the Hugging Face.

\subsubsection*{Dataset for Instruction Tuning}

To construct alignment data from the BIOMEDICA dataset, we transform image captions into instructional prompts. Following the methodology introduced in LLaVA-Med \cite{li2024llava}, we generate instructions based on the caption length:

\begin{enumerate}
    \item If the caption contains fewer than 30 words, we sample from a set of instructions prompting a brief description of the image.
    \item If the caption exceeds 30 words, we sample from a set of instructions prompting a detailed description of the image.
\end{enumerate}

\vspace{1em}
\subsection*{Model Development}

\subsubsection*{Multimodal Embedding Model Training}

We continuously pretrain OpenCLIP (ViT-L-14) \cite{ilharco_gabriel_2021_5143773} using 4 A100 GPUs for 36 epochs on a subset of 6M image-caption pairs from \dataset\ with an Information Noise-Contrastive Estimation (InfoNCE) loss \cite{oord2018representation}. To this end, we filtered \dataset, retaining only concepts within clinical and scientific imaging, immunoassays, illustrative diagrams, chemical structures, maps, tools and materials, and hand-drawn/screen-based visuals (thereby excluding tables, figures, and scientific equations). 

The InfoNCE loss is designed to maximize the similarity between correctly paired image and text embeddings while minimizing the similarity of incorrect pairs. Given \( N \) image-text pairs \( \{(\text{image}_1, \text{text}_1), \dots, (\text{image}_N, \text{text}_N)\} \) and their respective encoded representations:
 \(
z_{\text{image}_i }= E_{\text{image}}(\text{image}_i), \quad z_{\text{text}_i }= E_{\text{text}}(\text{text}_i), \quad \forall i \in \{1, \dots, N\},
 \)  we compute:

\begin{enumerate}
    \item Similarity Scores with Temperature Scaling:
    \[
    S_{k,j} = \frac{\text{sim}(z_{\text{image}_k}, z_{\text{text}_j})}{\tau}, \quad S'_{k,j} = \frac{\text{sim}(z_{\text{text}_k}, z_{\text{image}_j})}{\tau}
    \]
    
    \item Contrastive Loss for Image-to-Text Matching:
    \[
    \mathcal{L}_{I} = -\frac{1}{N} \sum_{k=1}^{N} \log \frac{\exp(S_{k,k})}{\sum_{j=1}^{N} \exp(S_{k,j})}
    \]
    
    \item Contrastive Loss for Text-to-Image Matching:
    \[
    \mathcal{L}_{T} = -\frac{1}{N} \sum_{k=1}^{N} \log \frac{\exp(S'_{k,k})}{\sum_{j=1}^{N} \exp(S'_{k,j})}
    \]
    
    \item Final InfoNCE Loss:
    \[
    \mathcal{L} = \frac{1}{2} (\mathcal{L}_{I} + \mathcal{L}_{T})
    \]
\end{enumerate}
\noindent where \( \text{sim}(\cdot, \cdot) \) denotes the cosine similarity between two embeddings, and \( \tau \) is a temperature parameter that controls the sharpness of the similarity distribution.


\subsubsection*{Multimodal Autoregressive Model Training}
We fine-tune SmolVLM \cite{marafioti2025smolvlm} using 2 A100 GPUs and Low-Rank Adaptation (LoRA) \cite{hu2022lora} via the Hugging Face SFT library with 10K instructions generated from \dataset. This framework fine-tunes a pretrained transformer while freezing most of its parameters, introducing trainable low-rank updates to selected layers. Given a sequence of tokens \( x = (x_1, x_2, ..., x_T) \), the model is optimized using the causal language modeling loss, defined as:  

\[
\mathcal{L} = -\sum_{t=1}^{T} \log p_{\theta}(x_t \mid x_{<t})
\]  

\noindent where \( p_{\theta}(x_t \mid x_{<t}) \) represents the predicted probability of token \( x_t \) given its preceding context.

\subsubsection*{Retrieval Augmented Generation}
We couple the \dataset\ index to a chain of LLMs, following a protocol similar to that of clinfo.ai \cite{lozano2023clinfo}.To this end, given a question and $N$ articles to collect (provided by user), our agent follows these  subsequent steps:
\begin{enumerate}
    \item Query generation: A question is parsed into a structured query using an LLM.
    \item Article retrieval: The generated query is used to retrieve $N$ relevant articles (ordered from most to least relevant) from the \dataset\ index.
    \item Evidence summarization: For each retrieved article, an LLM summarizes the evidence based on the full-text article as it pertains to the original question. If an article exceeds the context length, it is divided into smaller segments, and a multi-step refinement approach (e.g., summarizing each segment separately) is used for summarization.
    \item Final answer generation: The original question and the summarized evidence from $N$ articles are provided to an LLM to generate the final response.
\end{enumerate}

Unlike previous work, our approach allows the LLM to access full-text articles rather than just abstracts. While this provides richer information, it also introduces the challenge of processing a corpus often larger than those used in commercial LLMs. To address this, we employ a multi-step process, as the retrieved papers exceed the context window of any modern LLM.

\begin{figure}
    \centering
    \includegraphics[width=1\linewidth]{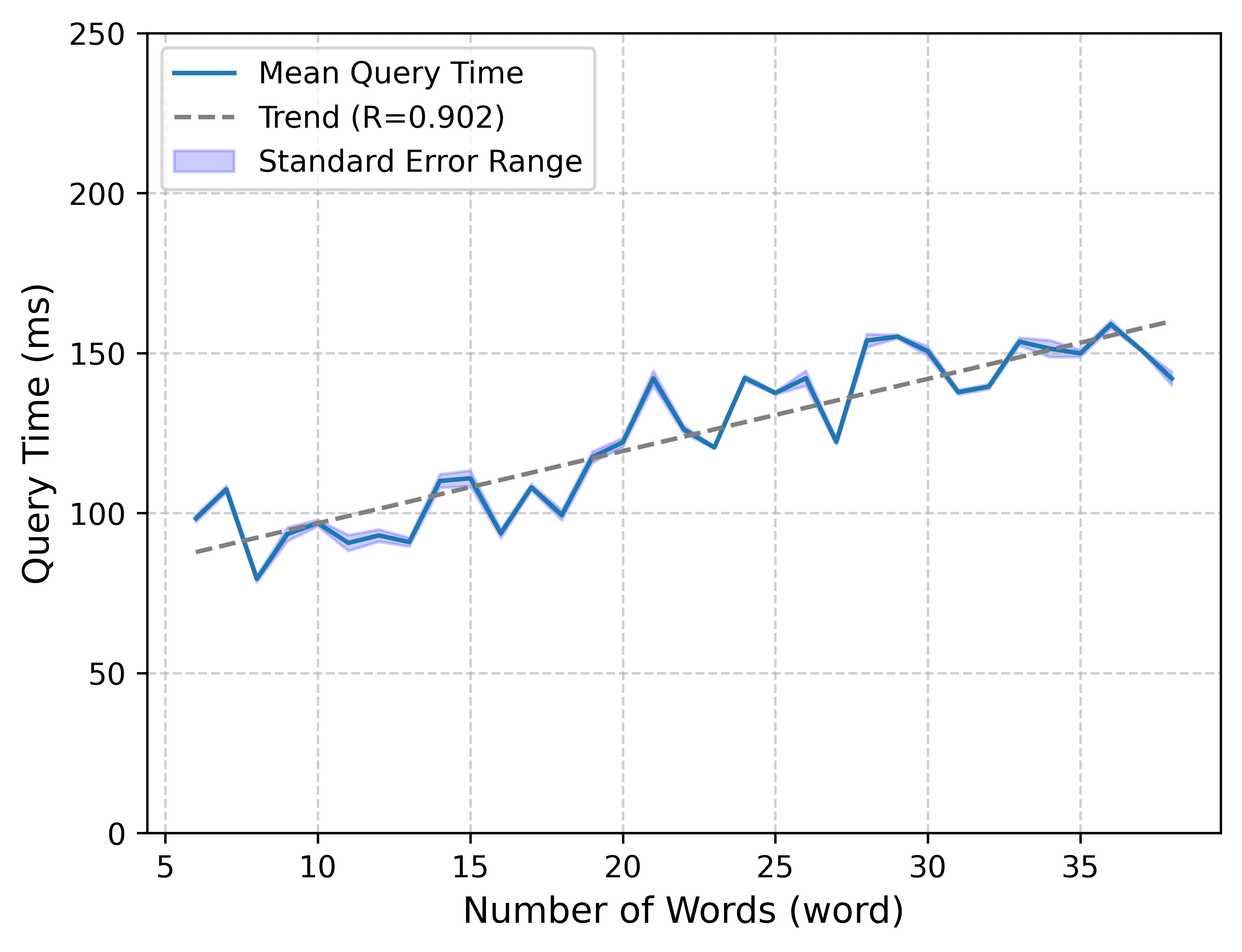}
    \caption{Query time as a function of the number of words (tokens) in the \dataset\ index. The blue solid line represents the mean query time, while the shaded blue region indicates the standard error. The dashed gray line denotes the linear trend in query time as token count increases $(R=0.902)$.}
    \label{query_time}
\end{figure}

\subsection*{Evaluation Benchmarks}

\begin{table*}[h!]
\small
\centering
\begin{adjustbox}{max width=\textwidth}
\begin{tabular}{lcccccc}
\toprule
 \textbf{Model} & \multicolumn{3}{c}{\textbf{Image $\rightarrow$ Text}} & \multicolumn{3}{c}{\textbf{Text $\rightarrow$ Image}} \\ 
\cmidrule(lr){2-4} \cmidrule(lr){5-7}
               & Recall@1 & Recall@10  & Recall@100 & Recall@1 & Recall@10  & Recall@100 \\ \hline

OpenCLIP \cite{ilharco_gabriel_2021_5143773}  &  2.78 & 9.78 &26.75 & 2.91 & 9.51 &24.59\\


\midrule

PMC-CLIP  \cite{lin2023pmc}   & 0.03 & 0.13 & 1.39 & 0.00 & 0.13 &1.50 \\   
BiomedCLIP  \cite{zhang2023biomedclip} & 3.70 & 12.78 & \underline{36.27} & \underline{3.94} & \underline{13.63} & \underline{35.63} \\

\midrule

\modelMed  &\textbf{ 4.13} &\textbf{ 15.13 }& \textbf{38.30 } & \textbf{4.15 }& \textbf{13.75} & \textbf{36.10}\\

\bottomrule
\end{tabular}
\end{adjustbox}
\caption{Top-K retrieval performance on BioMed-Flickr. \textbf{Bold} indicates best performance, \underline{underline} indicates second best performance.}
\label{table:retrieval_performance}
\end{table*}

\noindent \textbf{Image Classification Benchmark}
We construct a comprehensive classification benchmark by uniting evaluations from BioMedCLIP \cite{zhang2023biomedclip} and PMC-CLIP  \cite{eslami2023pubmedclip} and expanding underrepresented domains. Each task's classes are converted into two caption variations. The benchmark covers pathology (11), radiology (3), ophthalmology (1), dermatology (1), surgery (10), biology (9), and microscopy (4). Sources include Micro-Bench \cite{lozano2024mu} for biology/pathology, MedMNIST \cite{medmnst} for ophthalmology/dermatology, CheXpert  \cite{chexpert} for radiology, and the Dresden \cite{carstens2023dresden} dataset for M.I. surgery.

\noindent \textbf{Retrieval Benchmark}
We assess retrieval performance using a distinct set of 7K high-quality, open-source biomedical image-caption pairs from permissive image-caption pairs obtained from Flickr. This benchmark spans pathology, radiology, biology, dermatology, and surgery.

\noindent \textbf{Vision Question Answering}
We collect popular VQA datasets to evaluate autoregressive models including:  VQA-RAD \cite{lau2018dataset}, SLAKE \cite{liu2021slake}, PathVQA \cite{he2020pathvqa}, and PMC-VQA \cite{zhang2023pmc}.

\noindent \textbf{Medical Guidelines Dataset}
A double board-certified pathologist designed 50 challenging questions based on the latest guidelines in molecular pathology and pharmacogenomics. These questions focus on specific, recently updated medical  (within the training cut-off of current frontier models) knowledge rather than broad medical concepts (e.g. ``What is the most effective treatment for activated B cell-like (ABC) diffuse large B-Cell lymphoma (DLBCL) to achieve optimal outcomes?'')

\subsection*{Model Evaluation Protocol}

\subsubsection*{Embedding Model Evaluation}
We evaluate multimodal embedding models on zero-shot classification(formulated as closed VQA) and information retrieval benchmarks, as described in the evaluation benchmarks section. Each task has a corresponding protocol and evaluation metric:
\begin{enumerate}
  
   \item  \noindent \textbf{Image Classification to Closed VQA Formulation} Classification tasks are evaluated based on the average accuracy across two caption variations. All assessed multimodal embedding models include a vision encoder (\(E_{\text{image}}\)) and a text encoder (\(E_{\text{text}}\)). Given  a tuple of an  image and a list with  $M$ shuffled candidate answers (e.g. descriptions of the image), denoted as $(\mathbf{x}_i,[a_{i}^\pi(1),a_{i}^\pi(2),...,a_{i}^\pi(M)])$, we derive the image embedding, \(z_{x_i} = E_{\text{image}}(\mathbf{x}_i)\) alongside each candidate answer textual description: \(z_{a_{i}^\pi(j)} = E_{\text{text}}(a_{i}^\pi(j)) \hspace{0.5em} \forall \hspace{0.5em} j \in \{1,.., M_i\}\). Next, we calculate the cosine similarity score for each image-caption pair within a given tupple, \(s_{i,j} = z_{a_{i}^\pi(j)} \cdot z_{x_i}^T \hspace{0.5em} \forall \hspace{0.5em} j \in \{1,.., M_i\}\) to create the similarity vector $s = [s_{i,1},.., s_{i,M}]$. The element in $s$ with the highest  similarity  (\( \arg\max(s) \))  is selected as the final prediction. If the prediction  corresponds to the correct answer index, the question is considered correct; otherwise, incorrect.  

\vspace{1em}

  \item  \noindent \textbf{Information Retrieval Benchmark Formulation}

\noindent Given a dataset of $N$ samples, we evaluate retrieval performance using  Recall@k:  We first compute the image embedding, $z_{x_i}=E_{image}(\textbf{x}^i)$, along with  each caption in the dataset: $z_{c_i}=E_{text}( c^i )$ for $i \in \{1,.., M_i\}$. Then we compute the cosine similarity score for each caption, $s_{ij}=z_{c^i}\cdot z_{x_i}^T$ for $l \in[1,N_i]$. Captions are arranged from the largest to smallest similarity ($s_{ij}$). If the correct caption is within the first $k$-th arranged items, then the option is considered relevant, irrelevant otherwise. Lastly, we calculate Recall@k using the following equation:

\[
\text{Recall@k} = \frac{\text{Number of relevant items in the top } k \text{ results}}{\text{Total number of relevant items in the dataset}}
\]

\end{enumerate}

\vspace{1em}

 \subsubsection*{Autoregressive Model Evaluation}
\begin{enumerate}
    \item We evaluated multimodal autoregressive models on closed VQA tasks by measuring exact match  accuracy between the model's generated response and the ground-truth answer. Given an input image and corresponding question \( x_i,q_i \) and $M$ candidate answers \(a_i,...,a_M\) the model generates an answer \( \hat{a} \). Both predictions and answers are preprocessed (lowercased and punctuation removed).
    The prediction is considered correct if it exactly matches the reference answer \( a \). The overall accuracy is computed as:

\[
\text{Accuracy} = \frac{\text{Number of correctly predicted answers}}{\text{Total number of questions}}
\]

 \item For retrieval-augmented generation (RAG) models evaluated on zero-shot open QA, we used two clinicians as evaluators.  Given an input question  and retrieved context \( q_i ,C_i \), the model generates an answer \( \hat{a} \). The prediction is considered correct if it exactly matches the reference answer \( a \). The overall accuracy is computed.
\end{enumerate}

\section*{Data availability}
The dataset of image-caption pairs, metadata, and full-text article can be access via Hugging Face with just a few lines of code. We provide a step-by-step tutorial on how to access the dataset using Python in \url{bit.ly/biomedica-tutorial}.

\vspace{1em}

\section*{Competing interests}
The authors declare no competing interests.


\vspace{1em}
\noindent \textbf{Correspondence and requests for materials} should be addressed to
Alejandro Lozano, Min Woo Sun, and Serena Yeung-Levy.